\pdfoutput=1

\documentclass[11pt]{article}

\usepackage{acl}

\usepackage{times}
\usepackage{latexsym}

\usepackage[T1]{fontenc}

\usepackage[utf8]{inputenc}

\usepackage{microtype}

\usepackage{inconsolata}
\usepackage{graphicx}
\usepackage{url}
\usepackage{latexsym}
\usepackage{makecell}
\usepackage{epstopdf}
\usepackage{tipa}
\usepackage{hyperref}
\usepackage{xstring}
\usepackage{amsfonts}
\usepackage{amsmath}
\usepackage[T1]{fontenc}
\usepackage{booktabs}
\usepackage{arabtex}
\usepackage{hhline}
\usepackage{pdfpages}
\usepackage{amssymb}
\usepackage{utf8}
\newcommand{\hide}[1]{}

\newcommand{\AMADDA}{{\={A}}}
\newcommand{\AHAMZAUP}{{\^{A}}}
\newcommand{\WHAMZA}{{\^{w}}}
\newcommand{\AHAMZADN}{{\v{A}}}
\newcommand{\YHAMZA}{{\^{y}}}
\newcommand{\TAMARBUTA}{{$\hbar$}}

\newcommand{\THA}{{$\theta$}}
\newcommand{\DHA}{{\dh}}
\newcommand{\SHIN}{{\v{s}}}
\newcommand{\ZA}{{\v{D}}} 
\newcommand{\AYN}{{$\varsigma$}}
\newcommand{\GAYN}{{$\gamma$}}
\newcommand{\AMAQSURA}{{\'{y}}}

\newcommand{\FATHATAN}{{\~{a}}}
\newcommand{\KASRATAN}{{\~{\i}}}
\newcommand{\DAMMATAN}{{\~{u}}}
\newcommand{\SHADDA}{{$\sim$}}

\usepackage{multirow}
\usepackage{subfigure}
\usepackage{caption}
\setcode{utf8}
\vocalize
\usepackage[super]{nth}
\usepackage{tipa}
\usepackage{svg}
\usepackage{adjustbox}
\usepackage{acolor}

\usepackage{stackengine}
\usepackage{tikz}
\newcommand\dottedcircle{\tikz \draw [line cap=round, line width=0.15ex, dash pattern=on 0pt off 1.95\pgflinewidth] (0,0) circle [radius=0.6ex];}
\makeatletter
\newcommand{\fatha}{%
\bgroup \set@arabfont
\stackon[0.5pt]{\dottedcircle}{\char'013}%
\egroup%
}
\newcommand{\kasra}{%
\bgroup \set@arabfont
\stackunder[0.5pt]{\dottedcircle}{\char'013}%
\egroup%
}
\newcommand{\damma}{%
\bgroup \set@arabfont
\stackon[0.5pt]{\dottedcircle}{\char'014}%
\egroup%
}
\newcommand{\fathatan}{%
\bgroup \set@arabfont
\stackon[0.5pt]{\dottedcircle}{\char'023}%
\egroup%
}
\newcommand{\kasratan}{%
\bgroup \set@arabfont
\stackunder[0.5pt]{\dottedcircle}{\char'023}%
\egroup%
}
\newcommand{\shaddah}{%
\bgroup \set@arabfont
\stackon[0.5pt]{\dottedcircle}{\char'017}%
\egroup%
}
\newcommand{\dammatan}{%
\bgroup \set@arabfont
\stackon[0.5pt]{\dottedcircle}{\char'024}%
\egroup%
}
\newcommand{\skun}{%
\bgroup \set@arabfont
\stackon[0.5pt]{\dottedcircle}{\char'025}%
\egroup%
}
\newcommand{\alif}{%
\bgroup \set@arabfont
\stackon[0.5pt]{\dottedcircle}{\char'027}%
\egroup%
}
\makeatother

\newcommand{\widi}{{\bf WildDiacs}}
\newcommand{\MWM}{{\bf Wild2MaxDiacs}}
\newcommand{\wikinews}{{WikiNews}}
\newcommand{\wikinewsmax}{{WikiNewsMaxDiacs}}

\title{Arabic Diacritics in the Wild:\\ Exploiting Opportunities for Improved Diacritization}

\author{Salman Elgamal, Ossama Obeid, Tameem Kabbani,\textsuperscript{\textdagger} Go Inoue,\textsuperscript{$\ddagger$} Nizar Habash\\
  Computational Approaches to Modeling Language Lab, New York University Abu Dhabi\\
  \textsuperscript{\textdagger}American University of Sharjah\\
  \textsuperscript{$\ddagger$}Mohamed bin Zayed University of Artificial Intelligence\\
  \texttt{\{salman.elgamal,oobeid,nizar.habash\}@nyu.edu}\\
  \texttt{b00088948@aus.edu, go.inoue@mbzuai.ac.ae}
  }

\begin{document}
\maketitle
\begin{abstract}

The widespread absence of diacritical marks in Arabic text poses a significant challenge for Arabic natural language processing (NLP). This paper explores instances of naturally occurring diacritics, referred to as ``diacritics in the wild,'' to unveil patterns and latent information across six diverse genres: news articles, novels, children's books, poetry, political documents, and ChatGPT outputs. We present a new annotated dataset that maps real-world partially diacritized words to their maximal full diacritization in context. Additionally, we propose extensions to the analyze-and-disambiguate approach in Arabic NLP to leverage these diacritics, resulting in notable improvements. Our contributions encompass a thorough analysis, valuable datasets, and an extended diacritization algorithm. We release our code and datasets as open source.
\end{list}
\end{abstract}


\hide{

1. Intro

2. Arabic Diacritics
   Full/Partial
   Form/function

3. Related  
   * Data sets - ATB, Wikinews, Tashkila...
   * Role of D in Arabic NLP
   * Auto Diac
      >> mada approach vs. seq2seq

4. Diac in the wild
Statistics/pattern
  * 6 data sets with partial diac

5. Full Diac + Wellformedness
 * our data: 6 data sets with partial>> full diac
  * ATB + Wikinews  

6. Pilot Study...
Baseline: MorphAnalyzer >> Ranker

7. Exploiting Diacritics in the Wild

Approach: MorphAnalyzer >> Reranker >> ContextPostEdit

7.1 Reranker
7.2 MorphAnalyzer
7.3 ContextPostEdit

8. Evaluation
Results+Discussion.

}

\section{Introduction} 
Arabic orthography is infamous for its high degree of ambiguity due to its infrequently used optional diacritical marks.
While other Semitic languages like Hebrew and Syriac use similar systems, Arabic has a richer inflectional space with case endings and other orthographic choices that make Arabic more complex.
Interestingly, diacritical marks in Arabic are common in limited contexts where correct reading is a goal: holy texts, poetry and children's books, as well as books for adult literacy and non-native learners.
But in general reading contexts, for literate Arabic native speaker adults, diacritical marks are used frugally: $\sim$1-2\% of words are partially
diacritized
\cite{Habash:2010:introduction}.
We refer to these as Diacritics in the Wild 
({\widi}).

In this paper, we follow in the footsteps of other researchers to investigate whether such precious occurrences can be exploited to help improve the quality of Arabic NLP tools
\cite{diab-etal-2007-arabic,habash-etal-2016-exploiting,Bahar:2023:Take}.
We specifically focus on the Arabic diacritization task, which is at once  (a) a final target downstream application given the sparse and variable use of diacritics in Arabic; and (b) 
 an enabling technology for other applications such as speech synthesis \cite{halabi2016modern}.


While the percentage of {\widi} is small, our guiding intuitions are that on large scales, these are objects worthy of study, and given the extra information provided in such contexts, we assume the writers who added them wanted to provide hints to support optimal reading, e.g., to avoid garden path sentences. 

For a well-rounded exploration, we analyze and compare the diacritization patterns across six genres: news of multiple agencies, novels, children's books, poetry, political/legal documents of the UN, and ChatGPT output (which sometimes introduces diacritics unprompted).
Furthermore, we examine the diacritization patterns and choices  in two commonly used datasets for evaluating Arabic diacritization:  
The Penn Arabic Treebank \cite{Maamouri:2004:penn}
and {\wikinews} \cite{Darwish:2017:arabic-diac}.
We also develop a new annotated dataset that includes instances of partially diacritized words along with their \textit{full} diacritization, which we define carefully, acknowledging different practices. 
And finally, we propose an extension to the \textit{analyze-and-disambiguate} approach 
\cite{Habash:2005:arabic,Pasha:2014:madamira,Inoue:2022:morphosyntactic}
to improve the quality of its choices, and we evaluate on the data we annotated.



\begin{figure*}[t!]
\centering
\setlength{\tabcolsep}{3pt}
\begin{tabular}{c c}
(a) & (b) \\
\resizebox{0.7\textwidth}{!}{
\begin{tabular}{|c|c|c|c|c|c|c|c|c|c|c|c|c|}
\hline
\textbf{Fatha} & \textbf{Damma} & \textbf{Kasra} & \multicolumn{3}{c|}{\textbf{Nunation}} & \textbf{Shadda} & \textbf{Sukun} & \textbf{Dagger} & \multicolumn{4}{c|}{\textbf{Diacritic Clusters…}} \\ \hline
\Large{\<بَ>} & \Large{\<بُ>} & \Large{\<بِ>} & \Large{\<بً>} & \Large{\<بٌ>} & \Large{\<بٍ>} & \Large{\<بّ>} & \Large{\<بْ>} & \Large{\<بٰ>} & \Large{\<بُّ>} & \Large{\<بٌّ>} & \Large{\<بٍّ>} & \Large{\<بّٰ>} \\
\textit{ba} & \textit{bu} & \textit{bi} & \textit{bã} & \textit{bũ} & \textit{bĩ} & \textit{b$\sim$} & \textit{b.} & \textit{bá} & \textit{b$\sim$u} & \textit{b$\sim$ũ} & \textit{b$\sim$ĩ} & \textit{b$\sim$á} \\ \hline
\textit{ba} & \textit{bu} & \textit{bi} & \textit{ban} & \textit{bun} & \textit{bin} & \textit{bb} & \textit{b} & \textit{bā} & \textit{bbu} & \textit{bbun} & \textit{bbin} & \textit{bbā} \\
\hline
\end{tabular}
}
&
\begin{tabular}{c}
\vspace{-0.9em}\includegraphics[width=0.27\textwidth]{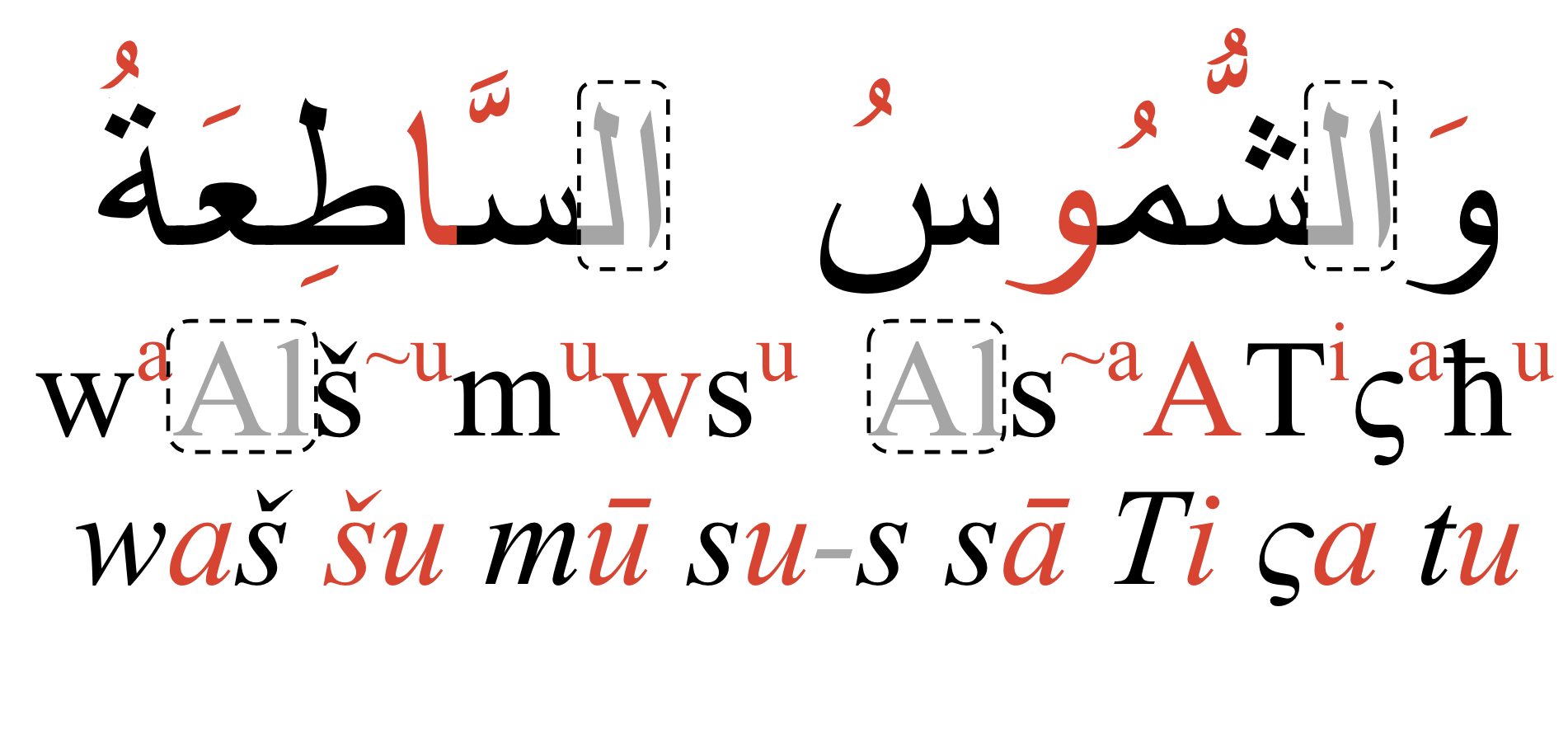}
\end{tabular}
\end{tabular}
\caption{(a) The nine Arabic diacritics commonly used in Modern Standard Arabic, grouped by function; and four examples of diacritic clusters. (b) A visually annotated example of a diacritized phrase meaning `and the bright suns [lit. and-the-suns the-bright]'. 
Diacritics are marked in red; and so are the undiacritized vowel-lengthening helping letters. Silent letters appear in dotted boxes. 
}
\label{fig:diac-list-example}
\end{figure*}

\textbf{Our contributions} are the following:
\begin{itemize}
\setlength{\itemsep}{0pt}%
\setlength{\topsep}{0pt} 
\setlength{\partopsep}{0pt}
\setlength{\parsep}{0pt}
\setlength{\parskip}{0pt}

\item We provide careful analysis and comparison of diacritization patterns in six genres of Arabic texts, shedding light on the needs and latent information in different Arabic genres.
\item We annotate a new dataset for studying maximal diacritization from partial diacritic signals ({\MWM}), and we extend an existing dataset, {\wikinews}, to address unhandled phenomena ({\wikinewsmax}).
\item We extend a hybrid (neuro-symbolic) algorithm for Arabic diacritization to make use of the existence of {\widi}, and demonstrate improved performance.  
\end{itemize}



Our code and datasets will be open-source and publicly available.\footnote{\url{https://github.com/CAMeL-Lab/wild_diacritics}}


The paper is organized as follows.
Section~\ref{diacritics} discusses Arabic diacritics.
Section~\ref{related} reviews related work.
Section~\ref{data} covers the datasets, including our annotated datasets, and {\widi} usage statistics.
Section~\ref{approach} details the approach we build on and how we extend it.
Section~\ref{eval} presents the experimental setup, evaluation, and error analysis.

\section{Arabic Diacritics} 
\label{diacritics}
\hide{



%

}

We present an overview of Arabic diacritics in terms of their form, function, and use.

\subsection{Arabic Diacritic Forms}
Arabic diacritics are zero-width characters that adorn Arabic letters 
to supplement Arabic's Abjad orthography with additional phonological signals.
There are many diacritical marks 
in the Arabic script: Unicode currently boasts 52 such symbols.\footnote{\url{https://unicode.org/charts/PDF/U0600.pdf}}
However, the basic \textit{Tashkil} (diacritization) set used in most Modern Standard Arabic (MSA) contexts includes nine symbols. See Figure~\ref{fig:diac-list-example}~(a).\footnote{
Arabic transliteration is in the HSB 
scheme~\cite{Habash:2007:arabic-transliteration}: 
  '~\<"'>,
  {\AMADDA}~\<آ>,
  {\AHAMZAUP}~\<أ>,
  {\WHAMZA}~\<ؤ>,
  {\AHAMZADN}~\<إ>,
  {\YHAMZA}~\<ئ>, 
A~\<ا>, b~\<ب>, {\TAMARBUTA}~\<ة>,
t~\<ت>, {\THA}~\<ث>, j~\<ج>, H~\<ح>, x~\<خ>, d~\<د>, {\DHA}~\<ذ>, r~\<ر>, z~\<ز>, s~\<س>, {\SHIN}~\<ش>, S~\<ص>, D~\<ض>, T~\<ط>, {\ZA}~\<ظ>, {\AYN}~\<ع>, {\GAYN}~\<غ>, f~\<ف>, q~\<ق>, k~\<ك>, l~\<ل>, m~\<م>, n~\<ن>, h~\<هـ>, w~\<و>, {\AMAQSURA}~\<ى>,  y~\<ي>,
  \textit{\FATHATAN}~\fathatan,
  \textit{\DAMMATAN}~\dammatan,
  \textit{\KASRATAN}~\kasratan,
  \textit{a}~\fatha,
  \textit{u}~\damma,
  \textit{i}~\kasra,
  \textit{\SHADDA}~\shaddah,
  \textit{.}~\skun,
  \textit{{\'a}}~\alif,
  {\"A}~\<ٱ>. }
%
In this paper, we focus on these MSA diacritics, which are all readily accessible by most Arabic keyboards.

%

{\textbf{Diacritic clusters}} can occur but are highly constrained.
The Shadda (\shaddah ~\textit{\SHADDA}) can combine with any one of the other diacritics, and none of them can combine with each other in MSA, except for  the Dagger Alif (\alif ~\textit{{\'a}}), which can follow Fatha (\fatha ~\textit{a}).
%
While in proper Arabic spelling the Shadda character should appear first in the string (and writing order) as it indicates doubling of previous consonant letter, it is important to note that a flipped order (e.g., Shadda after vowel) is impossible to detect visually.
We find both orders in the wild.
For example \textit{kat{\SHADDA}ab} and \textit{kata{\SHADDA}b} will always be rendered to appear as \textit{kat{\SHADDA}ab} (\<كَتَّب>).\footnote{Most text rendering libraries that support Arabic
implement the Arabic Mark Transient Reordering Algorithm (AMTRA) which normalizes the display of diacritic clusters \cite{pourander:2021:unicode}.
Furthermore, many systems utilize Unicode Normalization Forms (UNFs)~\cite{whistler:2023:unicode}
which, among other things, order adjacent diacritics based on their \textit{combining class} property
\cite{whistler:2023:unicodecharacter}.
UNFs are important for string matching and lexicographic sorting despite diacritic order variability.
These robustness-supporting mechanisms inadvertently allow inconsistent uses to coexist freely in the wild. \label{footnote-unf}
}



\hide{
}


\begin{table*}[t!]
    \resizebox{\textwidth}{!}{
    \begin{tabular}{|l|c|c|c|c|c|c|}
     \multicolumn{1}{c}{}& \multicolumn{1}{c}{(a)} & \multicolumn{1}{c}{(b)} & \multicolumn{1}{c}{(c)} & \multicolumn{1}{c}{(d)} & \multicolumn{1}{c}{(e)} & \multicolumn{1}{c}{(f)} \\ \hline
    \textbf{Arabic}
    & \<اليوم>
    & \<أشرقت>
    & \<الشمس>
    & \<الساطعة>
    & \<من>
    & \<الغرب>
    \\ \hline
    \textbf{English} & \textit{today} & \textit{rose} & \textit{the-sun} & \textit{the-bright} & \textit{from} & \textit{the-west} \\ \hline
    \textbf{Undiacritized} & \textit{Alywm} & \textit{Â{\SHIN}rqt} & \textit{Al{\SHIN}ms} & \textit{AlsAT{\AYN}{\TAMARBUTA}} & \textit{mn} & \textit{Al{\GAYN}rb}\\ \hline
    \textbf{Partial Diacritization} & \textit{Aly\textcolor{red}{\textbf{\underline{a}}}wm} & \textit{Â{\SHIN}r\textcolor{red}{\textbf{\underline{a}}}qt} & \textit{Al{\SHIN}ms\textcolor{red}{\textbf{\underline{u}}}} & \textit{AlsAT{\AYN}{\TAMARBUTA}} & \textit{m\textcolor{red}{\textbf{\underline{i}}}n} & \textit{Al{\GAYN}rb} \\ \hline
    \textbf{ATB Diacritization} & \textit{Aly\textbf{a}w\textcolor{red}{\textbf{\underline{.}}}m\textcolor{red}{\textbf{\underline{a}}}} & \textit{Â\textcolor{red}{\textbf{\underline{a}}}{\SHIN}\textcolor{red}{\textbf{\underline{.}}}r\textbf{a}q\textcolor{red}{\textbf{\underline{a}}}t} & \textit{Al{\SHIN}\textcolor{red}{\textbf{\underline{{\SHADDA}a}}}m\textcolor{red}{\textbf{\underline{.}}}s\textbf{u}} & \textit{Als\textcolor{red}{\textbf{\underline{{\SHADDA}}}}AT\textcolor{red}{\textbf{\underline{i}}}{\AYN}\textcolor{red}{\textbf{\underline{a}}}{\TAMARBUTA}\textcolor{red}{\textbf{\underline{u}}}} & \textit{m\textbf{i}n} & \textit{Al{\GAYN}\textcolor{red}{\textbf{\underline{a}}}r\textcolor{red}{\textbf{\underline{.}}}b\textcolor{red}{\textbf{\underline{i}}}} \\ \hline
    \textbf{WikiNews Diacritization} & \textit{Al\textcolor{red}{\textbf{\underline{.}}\textsuperscript{1}}y\textbf{a}w\textbf{.}m\textbf{a}} & \textit{Â\textbf{a}{\SHIN}\textbf{.}r\textbf{a}q\textbf{a}t\textcolor{red}{\textbf{\underline{.}}\textsuperscript{2}}} & \textit{Al{\SHIN}\textbf{{\SHADDA}}\textbf{a}m\textbf{.}s\textbf{u}} & \textit{Als\textbf{{\SHADDA}}\textcolor{red}{\textbf{\underline{a}}\textsuperscript{3}}AT\textbf{i}{\AYN}\textbf{a}{\TAMARBUTA}\textbf{u}} & \textit{m\textbf{i}n\textcolor{red}{\textbf{\underline{.}}\textsuperscript{2}}} & \textit{Al\textcolor{red}{\textbf{\underline{.}}\textsuperscript{1}}{\GAYN}\textbf{a}r\textbf{.}b\textbf{i}} \\ \hline
    \textbf{Maximal Diacritization} & \textit{A\textcolor{red}{\textbf{\underline{a}}\textsuperscript{4}}l\textbf{.}y\textbf{a}w\textbf{.}m\textbf{a}} & \textit{Â\textbf{a}{\SHIN}\textbf{.}r\textbf{a}q\textbf{a}t\textcolor{red}{\textbf{\underline{i}}\textsuperscript{5}}} & \textit{Al{\SHIN}{\SHADDA}\textbf{a}m\textbf{.}s\textbf{u}} & \textit{Als{\SHADDA}\textbf{a}AT\textbf{i}{\AYN}\textbf{a}{\TAMARBUTA}\textbf{u}} & \textit{m\textbf{i}n\textcolor{red}{\textbf{\underline{a}}\textsuperscript{5}}} & \textit{Al\textbf{.}{\GAYN}\textbf{a}r\textbf{.}b\textbf{i}} \\ \hline
    \textbf{Phonology} & \textit{‘al yaw ma} & \textit{‘a{\SHIN} ra qa ti-} & \textit{-{\SHIN} {\SHIN}am su-} & \textit{-s sā Ti {\AYN}a tu} & \textit{mi na-} & \textit{-l {\GAYN}ar bi} \\ \hline
    \end{tabular}
    }
    \captionof{figure}{An example in different levels of diacritization. The red underlined diacritics highlight changes from row to row. ATB is the Arabic Treebank diacritization standard \cite{Maamouri:2004:penn}, an essential full diacritization. {\wikinews} \cite{Darwish:2017:arabic-diac} addresses some ATB gaps like missing Sukuns (1,2) and long vowel marking with \textit{aA} (3). Maximal diacritization  adds contextual diacritics in word-initial (4) and inter-word contexts~(5).  }
    \label{fig:diac-types}
    \end{table*}

\subsection{Arabic Diacritic Functions}

\paragraph{Basic Phonological Mapping}
The diacritics primarily denote phonological information that supplements the Arabic Abjad orthography: vowel diacritics (Fatha, Damma, Kasra) indicate the presence of a short vowel; nunation (\<تنوين> \textit{tanwiyn}) diacritics indicate a short vowel followed by~\textipa{/n/};
the gemination diacritic, Shadda, indicates doubling of the consonant letter it follows; 
the Sukun (silence) diacritic indicates that no vowel is present; and finally, the special elongation diacritic (aka Dagger Alif)  indicates a long \textipa{/\=a/} vowel. 
See the red colored diacritics and their mapping to phonology in  Figure~\ref{fig:diac-list-example}~(b). 

An important and a not so obvious detail about the use of diacritics is that even under \textit{maximal diacritization}, some letters remain \textit{bare}, i.e., with no diacritics, to indicate specific phonological information. Examples include 
(i) \textit{Elongation}:
 the \textit{weak} letters 
(\<ا>~\textit{A},
\<و>~\textit{w},
\<ي>~\textit{y})
 remain bare when used as helping letters to elongate short vowel diacritics (Fig.~\ref{fig:diac-list-example}~(b)'s red letters); and
(ii) \textit{Silence}: letters that are phonologically elided or assimilated are marked by leaving them bare of diacritics (Fig.~\ref{fig:diac-list-example}~(b)'s grey letters).\footnote{A very common example is the \<ل> \textit{l} of the definite article \<ال> \textit{Al} when followed by a \textit{Sun Letter} -- a coronal consonant with which the \textit{l} assimilates, e.g., \textit{{\SHIN}} and \textit{s} in Fig.~\ref{fig:diac-list-example}(b).  
A Shadda on the following letter indicates assimilated gemination.}$^,$\footnote{The Alif (\<ا> \textit{A}) letter is used in word-initial positions as a vowel carrier (i.e., Alif Wasla, aka Hamzat Wasl).
When the vowel is elided in context, the Alif is retained but kept bare, e.g., cases of grey \<ا>~\textit{A} in Fig.~\ref{fig:diac-list-example}(b).  
Quranic Arabic retains the Wasla diacritic \<ٱ> \textit{\"{A}} to mark this absence, but MSA does not use it any more, which unavoidably leads to a minor ambiguity between silence and elongation.
We internally model the Alif Wasla as it interacts with other diacritization decisions, but convert it to a bare Alif in the output. See Appendices~\ref{appendix:morph-edits} and~\ref{appendix:contextual-edits}. 
}

\hide{
}

\paragraph{Dimensions of Disambiguation}
We can categorize the functional use of diacritics based on the disambiguating information they provide to the reader, which includes lexical, morphosyntactic, and contextual phonological liaison (\textit{sandhi}). 
%
%
For example, the word \<من>~\textit{mn} can be diacritized in different ways: 
\<مِنْ>~\textit{min.}~`from',
\<مَنْ>~\textit{man.}~`who', 
\<مَنَّ>~\textit{man{\SHADDA}a}~`he granted',
\<مَنٌّ>~\textit{man{\SHADDA}{\DAMMATAN}}~`a favor [indef. nom]', among many others. All four instances show lexical diacritics.
The last two words' final diacritics indicate morphosyntactic features such as verb aspect-person-gender-number, and nominal case and state.
In specific phonological contexts, some helping epenthetic vowels are introduced, and others may be elided. The typical epenthetic vowel is \kasra~\textit{i}, but there are other special cases:  
For example, \textit{min.} `from', has two additional forms: \textit{mina} before words starting with the definite article (see Fig.~\ref{fig:diac-types}), and  \textit{mini}  in general epenthesis, e.g., \<مِنِ ابْنِهِ> \textit{mini Ab.nihi} `from his son'.

\subsection{Arabic Diacritics in the Wild}
\label{sec:diacs-in-wild}
The two main challenges concerning Arabic diacritics in NLP are \textit{incompleteness} and \textit{inconsistency}.

\paragraph{The Challenge of Incompleteness}
Arabic diacritics are quite often omitted, with around 1-2\% of words in news text having at least one diacritic.
Arabic's rich templatic morphology, common default syntactic orders, and contextual semantic resolutions explain why educated Arabic readers
can read Arabic with such a signal deficit.
Not all texts and genres are equally devoid of diacritics.
For Quranic Arabic, and to a lesser extent poetry, diacritics are almost always included to avoid any misreading of the texts. 
Similarly for children's educational materials, diacritics are included to assist in learning. 
In this paper we study a number of genres to help understand how our approach will function under different conditions.

We define the terms \textbf{undiacritized} and \textbf{dediacritized} to refer to words with no diacritics, or stripped of all diacritics, respectively.
We will use the term \textbf{fully diacritized} to refer to a number of standards that have been used to evaluate automatic diacritization processes 
(Fig.~\ref{fig:diac-types} ATB and {\wikinews}).
We reserve the term  \textbf{maximally diacritized} to the version  we target in this paper as a higher standard of completeness (Appendix~\ref{appendix:fulldiacs}).
\textbf{Partially diacritized} refers to a state of in-betweenness where some diacritics are provided. {\widi} are typically partial diacritizations.

\paragraph{The Challenge of Inconsistency}

We note two types of inconsistency in diacritic use.
First, there are a number of acceptable variations that reflect different styles. 
These appear in the wild, as well as in commonly used evaluation datasets.
Examples include (i) foreign names whose diacrtization reflect local pronunciation such as \<بريطانيا> `Britain' as \textit{b.riTaAniyaA} or \textit{biriTaAniyaA}; (ii) the inclusion of the Sukun (silence diacritic) at the end of utterance-final words; and (iii) the position of the Tanwiyn Fath before or after a word-final silent Alif or Alif-Maqsura such \<كِتَابًا>  \textit{kitaAb{\FATHATAN}A} or \<كِتَاباً>  \textit{kitaAbA{\FATHATAN}} `a book'.

The second type are simply errors. Examples include (i) the Shadda appearing after the vowel diacritic, e.g., \textit{kata{\SHADDA}b} instead of the correct \textit{kat{\SHADDA}ab} `he dictated'; (ii) multiple incompatible diacritics on the same letter, e.g., \textit{ktAbuu} instead of \textit{ktAb{\DAMMATAN}}; (iii) diacritics in impossible positions such as word initial \textit{{\KASRATAN}ktAb}; and (iv) the correct diacritic is on the incorrect letter, e.g., \textit{ktiAb} for \textit{kitAb}.

To catch some of these inconsistencies, we developed a well-formedness check script and used it as part of this paper. Details are in Appendix~\ref{appendix:wellfornmedness}.

Despite their imperfections, {\widi} are useful human annotations that not only aid other human readers, but can be exploited automatically.
%
%

\hide{
In the wild, there is no mechanism to enforce this ordering constraint, and in fact, we find many nonsensical combinations do occur (Section~\ref{data}).

Another design decision we took was enforcing Tanwiyn Fath to come before Alif and Alif Maqsura (\<ٮًا> $\circ$FA and \<ٮًى> $\circ$FY), instead of the equally ubundant alternatives ($\circ$AF and $\circ$YF) (Section \ref{data}). This simplifies our full diacritization rules due to the reduction in the number of different letters that can appear bare.

We take contexts as separated by punctuation marks and sentence boundaries. We finally include contextual well-formedness as the last requirement for our definition of full diacritization well-formedness.

We then define the rules of Arabic phonology we enforce for a full diacritization to be considered well-formed (these rules exclusively govern diacritics, in other words, we assume that a morphologically valid lexeme is given to start with):
 
}

\section{Related Work} \label{related} 

\subsection{Roles of Diacritics in Arabic NLP}
Diacritics play a significant role in a wide variety of NLP tasks, such as text-to-speech synthesis~\cite{ungurean:2008:automatic,halabi2016modern}, automatic speech recognition~\cite{aldarmaki:2023:diacritic}, machine translation~\cite{diab-etal-2007-arabic,alqahtani-etal-2016-investigating,fadel-etal-2019-neural,thompson-alshehri-2022-improving}, morphological annotation~\cite{habash-etal-2016-exploiting}, homograph disambiguation~\cite{alqahtani-etal-2019-homograph}, linking lemmas across different lexical resources~\cite{jarrar2018diacritic}, language proficiency assessment~\cite{hamed-zesch-2018-role}, and improving text
readability~\cite{esmail-etal-2022-much,Elnokrashy:2024:partial}.

Several attempts have explored the impact of varying degrees of diacritization in downstream tasks.
\citet{diab-etal-2007-arabic} and their follow-up work~\cite{alqahtani-etal-2016-investigating} investigate the impact of various degrees of diacritization to identify the optimal amount of diacritics for better machine translation performance.
\citet{habash-etal-2016-exploiting} observe a positive correlation between the degree of diacritization in the input text and the performance in the morphological annotation task.

Using naturally occurring diacritics in the input text is shown to be effective in the diacritization task itself.
\citet{alkhamissi-etal-2020-deep} propose a model that accepts partially diacritized text during decoding, demonstrating improved diacritization performance.
With a similar motivation, \citet{Bahar:2023:Take} introduce a bi-source model that takes both characters and optional diacritics available in the input sequence.
Our work differs from theirs in that no training is required to leverage the presence of diacritics in the input text, making our approach computationally efficient.

\subsection{Datasets for Diacritization}
Numerous datasets have been developed for diacritization in different variants of Arabic, such as MSA~\cite{Maamouri:2004:penn, Darwish:2017:arabic-diac}, classical Arabic~\cite{Zerrouki:2017:tashkeela,yousef2019learning,fadel2019arabic}, and dialectal Arabic~\cite{Jarrar:2016:curras,abdelali:2019:diacritization,el-haj-2020-habibi, alabbasi-etal-2022-gulf}.
The source of datasets varies, including news, e.g., the Penn Arabic Treebank (ATB)~\cite{Maamouri:2004:penn} and the {\wikinews} dataset~\cite{Darwish:2017:arabic-diac}, classical books~\citep[Tashkeela;][]{Zerrouki:2017:tashkeela}, and poetry~\citep[APCD;][]{yousef2019learning}.
Among these resources, ATB and {\wikinews} are widely used as the standard benchmark datasets for diacritization in MSA.

Annotation conventions vary across datasets due to the lack of consensus in definitions, and may even be inconsistent within a dataset~\cite{Darwish:2017:arabic-diac}.
In this work, we thoroughly analyze and compare diacritization patterns in widely used datasets, highlighting the need for an evaluation set based on \textit{maximal diacritization}---a refined definition of diacritization.
We also provide a new annotated dataset based on this paradigm comprising 6,000 words across six different genres.

\subsection{Automatic Diacritization}
Approaches to automatic diacritization vary in task formulation, architecture choice, and the use of external resources.
One line of work formulates diacritization as a single isolated task, e.g., a character-level sequence labeling problem~\cite[among others]{zitouni-etal-2006-maximum} and a sequence-to-sequence problem~\cite{mubarak-etal-2019-highly}.
Early efforts employ traditional machine learning models such as the maximum entropy classifier~\cite{zitouni-etal-2006-maximum} and SVMs~\cite{Darwish:2017:arabic-diac}.
Recently, neural models have shown significant progress, such as LSTMs and their extensions~\cite{Abandah:2015:automatic, belinkov-glass-2015-Arabic, fadel-etal-2019-neural, madhfar:2021:effective, darwish:2021:arabic,elmallah-etal-2024-arabic-diacritization} and Transformer-based models~\cite{mubarak-etal-2019-highly,qin-etal-2021-improving,elmadany-etal-2023-octopus}.

Another line of work addresses diacritization within a multi-task setup, jointly modeling diacritization with relevant tasks such as POS tagging~\cite[among others]{Habash:2005:arabic,alqahtani-etal-2020-multitask} and machine translation~\cite{thompson-alshehri-2022-improving}.
A large body of this kind has demonstrated the value of using external resources such as a morphological analyzer.
They adopt an \textit{analyze-and-disambiguate} strategy, where they generate possible analyses for each word with a morphological analyzer (\textit{analyze}), then rank the analyses based on separately trained feature classifiers (\textit{disambiguate}).
This approach has been extensively explored using various architectures, including SVMs~\cite{Habash:2005:arabic,habash-rambow-2007-arabic,roth-etal-2008-arabic,Pasha:2014:madamira,shahrour-etal-2015-improving}, LSTMs~\cite{Zalmout:2017:dont,zalmout-habash-2019-adversarial,zalmout-habash-2020-joint}, and Transformer-based models~\cite{Inoue:2022:morphosyntactic,Obeid:2022:camelira}.
In this work, we present an extension to this approach where we utilize the presence of naturally occurring diacritics in the input text to improve the re-ranking of the analyses without additional training.

\begin{table*}[t!]
    \centering
    \includegraphics[width=0.97\textwidth]{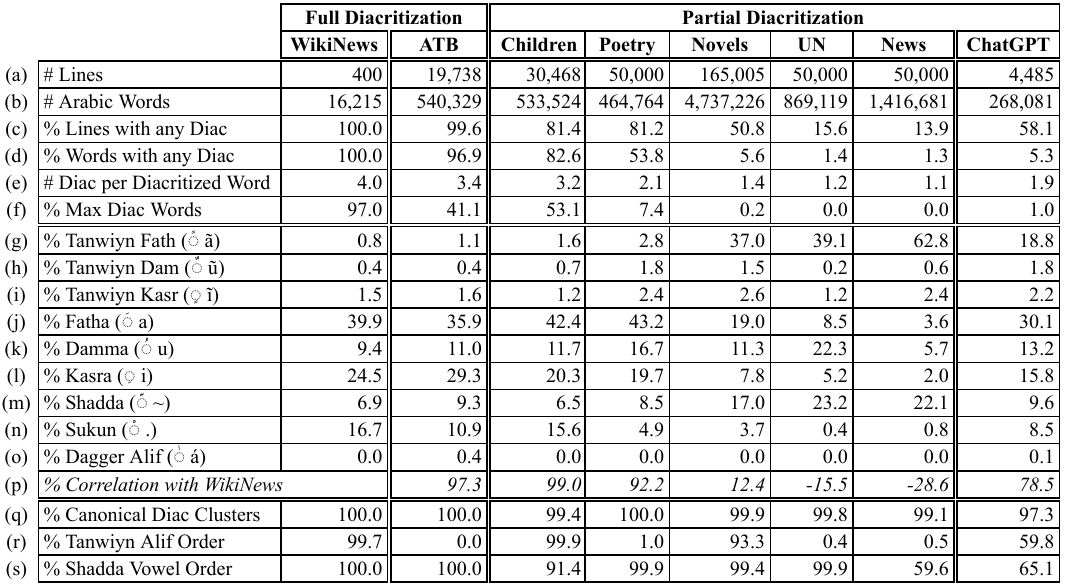}

    \caption{Statistics of diacritic usage in fully diacritized and partially diacritized datasets.}
    \label{tab:diac-statistics}
\end{table*}

\section{Diacritics: Data and Statistics} 
\label{data}

\subsection{Datasets}
We use eight pre-existing datasets (two fully diacritized and six partially diacritized), 
and we introduce two new maximally diacritized datasets that we  make publicly available.

\paragraph{Fully Diacritized Datasets}
We include the {\wikinews}~\cite{Darwish:2017:arabic-diac} and ATB~\cite{Maamouri:2004:penn} datasets, which have been used extensively in past recording on Arabic diacritization \cite{Darwish:2017:arabic-diac,Pasha:2014:madamira,mubarak-etal-2019-highly}.  As mentioned in Sec.~\ref{sec:diacs-in-wild} and Fig.~\ref{fig:diac-types}, {\wikinews}' annotation guideline lead to a fuller diacritization than those of ATB's essential guidelines.  In Table~\ref{tab:diac-statistics}, we use all the text of {\wikinews}, and the \textit{training} data portion of ATB Parts~1-3 \cite{Diab:2010:guidelines}.

\paragraph{Partially Diacritized Datasets}
To study the variation in {\widi} patterns in different genres, we targeted six datasets covering News, Poetry, Novels, Children's books, UN, and ChatGPT. 
%
The \textbf{Children} and \textbf{Novels} set are from the Hindawi website\footnote{\url{https://www.hindawi.org/}} \cite{AlKhalil:2018:leveled}.  The \textbf{Poetry} samples are from the Arabic Poem Comprehensive Dataset (APCD) \cite{yousef2019learning}. 
The \textbf{UN} data is from the Arabic portion of the UN Corpus \cite{Ziemski:2016:united}.
The \textbf{News} data is from Arabic Gigaword \cite{Parker:2011:arabic}, specifically from Asharq Alawsat (aaw) and AlHayat (hyt). We intentionally did not use the sources used in building the ATB corpus (afp, umh, nhr, asb) since ATB was used to train the baseline model we used in our experiments, and we did not chose Ahram (ahr) or Xinhua (xin) because of their very low percentage of {\widi}.  Finally, for \textbf{ChatGPT}, we use the dataset released by \newcite{alhafni-etal-2023-advancements} as part of their work on grammatical error correction, where they prompted ChatGPT-3.5 to correct undiacritized raw Arabic text and ChatGPT interestingly added seemingly random partial diacritizations.  
We randomly sampled 50,000 lines from Poetry, UN, and News due to their large size, while we used the whole datasets for the rest of the genres.%
\footnote{All texts are processed with CAMeL Tools' simple word tokenizer \cite{Obeid:2020:camel}.}

\paragraph{Maximally Diacritized Datasets}
We introduce two new datasets. 
The first is the \textbf{Multi-genre Wild Diacritization to Maximal Diacritization} ({\MWM}) dataset.  
{\MWM} is composed of randomly chosen 6,000 words with wild partial diacritization, 1,000 from each of the six genres discussed above. 
For each selected word, given its full context,
 we provide a maximally diacritized version of it.  The annotations were carried out by two Arabic native speakers. 
The annotators closely followed our definition of in-context full diacritization well-formedness. The annotations were passed through our well-formedness checks as an extra pass of validation and the words that failed our check were corrected. 
{\MWM} is split into equal 3,000-word development (Dev) and Test sets.

The second dataset is \textbf{{\wikinewsmax}}, a new version of {\wikinews} that we manually extended to our maximal diacritization, affecting 3.5\% of its tokens. The main extensions are adding contextual diacritics and Dagger Alifs.
We also, for the first time to our knowledge, annotated the dataset with valid alternative diacritizations (2.1\% of tokens), such in the case of foreign names discussed above.

\subsection{Statistics} 
Table~\ref{tab:diac-statistics} presents different diacritic usage statistics.
%
\paragraph{Diacritization Completeness Across Genres}
Table~\ref{tab:diac-statistics}~(c-d) highlight the great variation in use of diacritics across different genres.
The fully diacritized datasets have almost complete diacritization by design.
The columns in Table~\ref{tab:diac-statistics} are ordered to reflect the pattern of completeness (from left to right), except that ChatGPT, being artificial, is separated at the end.
As expected, the degree of completeness in naturally occurring partially diacritized datasets is highest for \textbf{Children} text, followed by \textbf{Poetry}, \textbf{Novels}, \textbf{UN} documents and finally \textbf{News}.
\textbf{ChatGPT} has an interestingly high level of {\widi} comparable to Novels.
We also observe that the number of diacritics per diacritized word correlates highly with the percentage of words with diacritics and the percentage of fully diacritized words---Table~\ref{tab:diac-statistics}~(d-f).  The low number of fully diacritized words for ATB and lower than perfect for {\wikinews} reflect the systematically missing diacritics in their annotations (see Section~\ref{sec:diacs-in-wild}).\footnote{The main issues in ATB diacritization are (i) no \textit{a} diacritic before Alif when indicating long vowel \textipa{/\=a/} , (ii) no Sukun to mark the definite article with moon letters and some foreign name consonant clusters, and (iii) no contextual diacritics. \label{ATB-footnote}}


\paragraph{Diacritization Patterns Across Genres}

Table~\ref{tab:diac-statistics}~(g-o) shows the distributions of the nine diacritics (in Unicode order) across the datasets relative to the total number diacritics.  
%
Table~\ref{tab:diac-statistics}~(p) presents the correlation of the diacritic distributions of each genre against the {\wikinews} distribution (fullest in diacritization).
Interestingly, we note that the genres with high percentages of diacritics are highly correlated with full diacritization; but as percentages drop, the ratio of lower frequency diacritics are more common (and the correlation becomes negative).
This makes sense as it suggests {\widi} fill in the gaps for the low frequency events.	This is most evident when comparing \textbf{News} to \textbf{{\wikinews}} (same genre but different degrees of completeness): 
Sukuns drop in News to 1/20  of their {\wikinews} distribution;
and Tanwiyn Fath increases by over 70 times.
We also note that the ChatGPT correlation pattern with {\wikinews} is oddly different from all naturally occurring partial diacritization datasets.
Finally, we  note that the use of Dagger Alif is almost extinct across MSA; and it seems to mostly be present in the ATB tokenization. This is the only case where ATB has more details than {\wikinews}. We include Dagger Alif in our Maximal Diacritization.

\paragraph{Wild Diacritization Failures}
Table~\ref{tab:diac-statistics}~(q) reports that almost all of the  diacritized words pass our well-formedness check, with ChatGPT performing the worst by far.
Table~\ref{tab:diac-statistics}~(r) shows that the Alif-Tanwiyn order is split between two schools: \textit{{\FATHATAN}A} in {\wikinews}, Children and Novels; and \textit{A{\FATHATAN}} in ATB, Poetry, UN and News. ChatGPT produces a mix of the two approaches.  
And finally, according to Table~\ref{tab:diac-statistics}~(s) the Shadda-Vowel order is almost perfectly stable, except in News and ChatGPT, where there is a lot of noise.

\hide{
Table \ref{tab:diac-statistics} presents different statistics in the behaviour of WiDi in the our six datasets. The rows are divided into sections: rows 1-2 show line specific data, namely, the number of lines analysed in each dataset and the percentage of lines with at least one diacritic. Rows 2-5 present word specific data, namely, the percentage of Arabic words with with at least one diacritic, the average number of diacritics per diac word (word with at least one diacritic) and the percentage of fully diacritized arabic words according to our well-formedness definition of full diacritizatoin (Section \ref{diacritics}). Rows 6-14 present the relative occurance percentages of diacritics. Rows 15-17 present the relative occurance of different forms of some diacritical patterns.

It is worth noting that the datasets vary in the percentage of lines with WiDis (row 2), percentage of arabic words with WiDis (row 3) and the average frequency of WiDis per partially diacritized arabic word (row 4).
All three quantities follow similar trends with {\wikinews} having the highest frequency of WiDis, then ATB, Children, Poetry, in that specific order, and then at similar WiDi frequency levels come Novels and ChatGPT, and then lastly UN. This information is consistent with the importance of diacritics to the purpose of the dataset.

Based on our definition of full diacritization well-formedness, row 5 of table \ref{fig:diac-statistics} shows that {\wikinews} scored the highest percentage of well-fromedness at 97.0\% of arabic words, while the other reference dataset, ATB, only scored 41.1\%.
The other datasets scored significantly lower percentages since they are taken from texts that isn't intended to be always fully diacritized, with the exception of Children which scored 53.1\% due to it's educative purpose and Poetry at 7.4\% since it has a higher frequency of WiDis in general.

We then further analyse the words that failed the well-formedness check.
Table \ref{fig:ref-well-formedness-statistics} shows the percentage frequencies of different error types out of the total number of ill-formed words.
Note that these categories are not mutually exclusive, with the exception of the last \textit{Uncategorized} category which includes the ill-formed words that can not be place under one or multiple of the remaining error categories.

The five types of errors analysed in table \ref{fig:ref-well-formedness-statistics} are:
\begin{itemize}
    \setlength{\itemsep}{0pt}%
    \setlength{\topsep}{0pt} 
    \setlength{\partopsep}{0pt}
    \setlength{\parsep}{0pt}
    \setlength{\parskip}{0pt}

    \item Shadda Error: Mis-order of Shadda in a diac cluster, i.e. Shadda not placed at the beginning of its cluster.
    \item Tanwiyn Error: Mis-order of Tanwiyn Fath, i.e. using ($\circ$AF and $\circ$YF) instead of ($\circ$FA and $\circ$FY).
    \item Moon Lam Error: Missing the sukun on the lunar (\<ل> l) in the definite article (\<ال> Al)\footnote{Only at the beginning of words due to the ambiguity of the definite article when preceded by a prefix}.
    \item Bare Alif Error: Missing short vowel diacritic before Alif (\<ا> A)\footnote{With the exception of Alif Fariqa and Alif Tanwiyn Fath}.
    \item Contextual Error: Error in contextual diacritics.
\end{itemize}

Table \ref{fig:ref-well-formedness-statistics} then shows that the vast majority, 96.7\%, of ill-formed words in {\wikinews} involve contextual errors, since {\wikinews} doesn't handle contextual diacritics.
Alternatively, ATB includes a large variety of errors with the most frequent being Bare Alif Errors at 59.4\% and Moon Lam Errors at 29.6\%, while contextual errors are at 4.26\% of all ill-formed words yet still standing at 2.51\% of all arabic words in ATB.
ATB also includes a higher percentage of \textit{Uncategorized} errors which includes letters with missing diacritics that are harder to draw error patterns from.

Table \ref{fig:general-statistics} then includes the relative distribution of diacritical marks in rows 6-14.
It is worth highlighting that the Fatha and Tanwiyn Fath are always the most frequently used diacritic in every dataset, with Fatha being the most frequent in ATB, {\wikinews}, Poetry, Children, and ChatGPT and Tanwiyn Fath being the most frequent diacritic in Novels, UN and News despite arguably being insignificant in increasing text clarity for readers.

Finally, rows 15-17 of the table show that diac cluster canonicality is preserved to a high extent (> 99\%) accross all datasets, except for ChatGPT which has a canonicality level of 97.3\% in its diacritic clusters, with 98.8\% of those ChatGPT non-canonical clusters are Shadda Errors.
Tanwiyn Fath's order also appears to from dataset to dataset, with {\wikinews}, Children, Novels choosing primarily to put Tanwiyn Fath before Alif and Alif Maqsura, consistent with our well-formedness definition, while ATB, Poetry, UN and News are primarily placing Tanwiyn Fath after Alif and Alif Maqsura, while ChatGPT inconsistently includes both orders while being slightly biased towards our well-formed order.

\begin{figure*}[t!]
    \centering
    \includegraphics[width=1.0\textwidth]{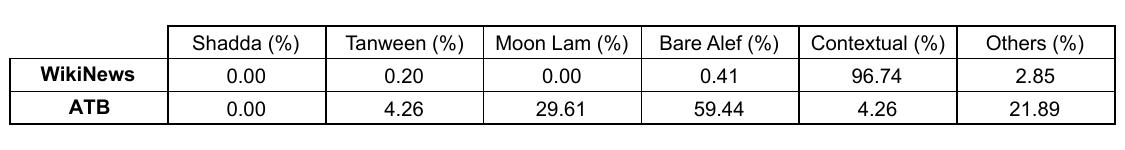}
    \caption{Analysis of Ill-formed Words in Reference Datasets}
    \label{fig:ref-well-formedness-statistics}
\end{figure*}

We then further analyse the words that failed the well-formedness check.

Table \ref{fig:ref-well-formedness-statistics} shows the percentage frequencies of different error types out of the total number of ill-formed words. Note that these categories are not mutually exclusive, with the exception of the last \textit{Uncategorized} category which includes the ill-formed words that can not be place under one or multiple of the remaining error categories.

}

\begin{table*}[t!]
    \centering
    \includegraphics[width=0.55\textwidth]{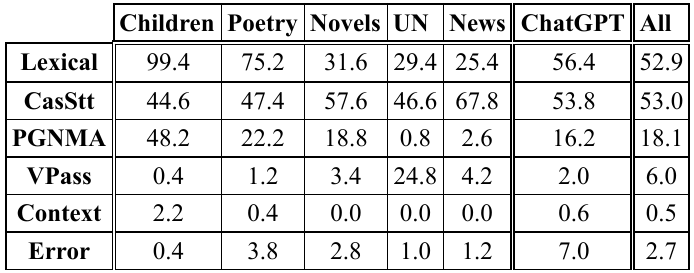}

    \caption{Percentage of words where {\widi} 
    indicate Lexical specification, CasStt (case, state), PGNMA (person, gender, number, mood, aspect), VPass (passive voice), and contextual diacritics, or are errors. }
    \label{tab:diac-func-statistics}
\end{table*}

\paragraph{Functional Use of Wild Diacritization}
\begin{figure*}[t!]
    \centering
    \begin{center}
       \includegraphics[width=0.9\textwidth]{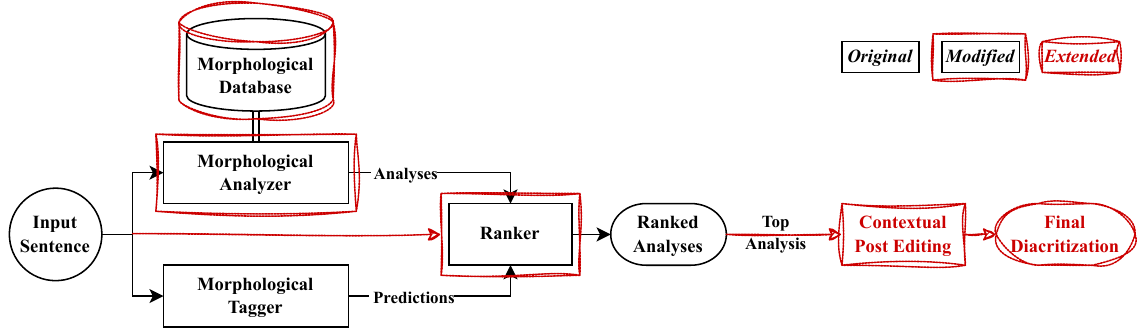} 
    \end{center}
    \caption{The original morphological analysis and disambiguation process with modifications and extensions.}
    \label{fig:ct-architecture}
\end{figure*}

Finally, we report on the distribution of functional uses of {\widi} in Table~\ref{tab:diac-func-statistics}.
We annotated the 3,000 words in the {\MWM} Dev set across all six genres for six categories: lexical, case/state, person/gender/number/mood/aspect (PGNMA), passive verb, context diacritics, and errors.
Lexical and case/state are the highest in frequency overall, followed by PGNMA then voice.
The most dominant genres per category are: lexical, PGNMA and context (Children), case/state (News, 94\% Tanwiyn Fath), passive voice (UN), and Errors (ChatGPT).
The diversity and patterns of functional use focus are consistent with our expectations, although not previously reported to our knowledge.

\section{Exploiting Diacritics in the Wild} 

\begin{table*}[t!]
    \centering
  \includegraphics[width=0.80\textwidth]{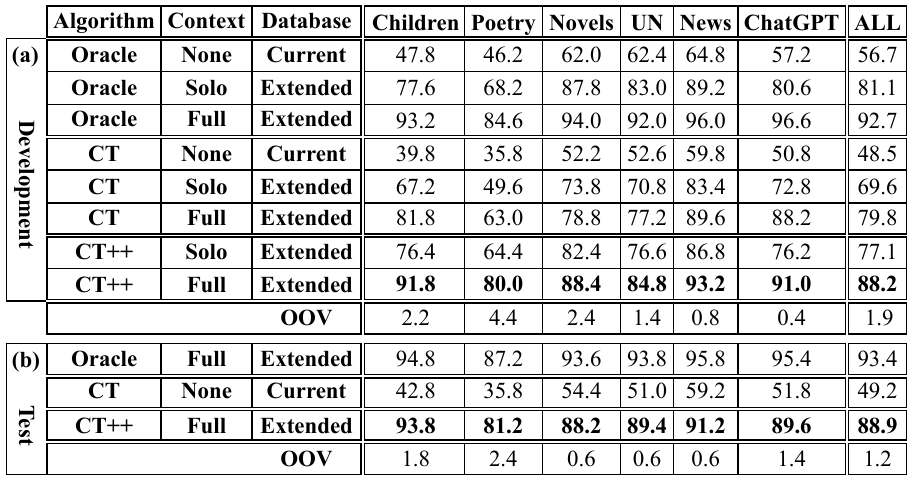}
    \caption{Percentage of correctly maximally diacritized words from {\MWM} dataset: (a) Dev and (b) Test. For Dev, all possible system combinations are presented, except for impossible combinations, e.g., the Current DB with the Context Full is meaningless since the Context Full requires information not in the Current DB. For Test, only the best setup identified by the Dev set, baseline and oracle are presented.}
    \label{tab:dev_results}
\end{table*}

\label{approach}
Our basic approach to exploiting {\widi} is based on the intuition that they are simply \textit{hints} that we need to \textit{take} \cite{Bahar:2023:Take}.  
We build on a hybrid (neuro-symbolic) \textit{analyze-and-disambiguate} 
 solution for Arabic diacritization 
 implemented in CAMeL Tools \cite{Obeid:2020:camel}.
While this open-source toolkit 
reports competitive results \cite{Obeid:2020:camel,Obeid:2022:camelira}, it has a number of limitations connected to its dependence on the-less-than-ideal ATB diacritization (training and tools).
We present the basic approach followed by our modifications and extensions.  

\subsection{The Analysis \& Disambiguation Process}
Figure \ref{fig:ct-architecture} shows the basic  CAMeL Tools disambiguation process (in black components) and our extensions and modifications of it (in red components).
First, for each word in the input sentence, a list of analyses is generated by a symbolic morphological analyzer.
These analyses are comprised of a number of morphological features (e.g., part-of-speech, gender, case), a number of lexical features (e.g., lemma, diacritized orthography), and a number of pre-computed statistical features (e.g., log probability of lemma).
In parallel, a neural morphological tagger predicts a set of morphological features (a subset of features provided in analysis) for each word in context.\footnote{We report here on the Unfactored BERT Disambiguator model \cite{Inoue:2022:morphosyntactic}.}
Finally, the analyses are ranked using the tagger's predictions
by sorting on the number of matching morphological features per analysis, then breaking analysis ties with the following analysis features in order: the joint POS-lemma log probability, the lemma log probability, and finally sorting lexicographically on the analysis diacritization, to ensure a final stable ranking.

\subsection{Modifications \& Extensions}

\paragraph{Re-ranking with {\widi}}
We modified the basic ranker to use Levenshtein insertion, deletion, and substitution edits between input word and analysis diacritization.
Empirically, our best setup is sorted with substitutions+deletions, followed by morphological features, substitution alone, deletion alone, POS+lemma log probability, lemma log probability, insertions, then diacritization lexicographic order.
This order guarantees ideal behavior when the input is fully diacritized, and is backward compatible with the original ranker when the input is not diacritized. See more details in Appendix~\ref{appendix:ranker}.

\paragraph{Morphological Analyzer Modifications}
Our approach depends heavily on the choices produced by the morphological analyzers.
The baseline analyzer we use is based on SAMA/Calima-Star \cite{Graff:2009:standard,taji-etal-2018-arabic}
which follow the ATB full diacritization standard.
As such, we make a number of changes to the morphological analyzer's algorithm and database to accommodate maximal diacritization (Footnote~\ref{ATB-footnote}).
Examples include (i) adding a missing Fatha before Alif for long vowel representation, (ii) fixing the position in A/{\AMAQSURA} word-final Tanwiyn Fath, (iii) adding missing Sukuns in word final and medial locations,  
(iv) enhancing the database's Alif Wasla coverage to include unmodeled cases, e.g., the determiner \<ٱل>~{\textit{\"Aal}},
and (v) adding string flags to mark cases with allomorphic variants, e.g., \<مِن> \textit{min} is marked as \textit{min\%n} (Section~\ref{sec:diacs-in-wild}). 
For more details, see Appendix~\ref{appendix:morph-edits}.




\paragraph{Contextual Post Edit Extensions}
After analysis re-ranking, the top analysis for each token is selected, and an array of all top analyses are  passed through a new component that handles contextual post edits. 
The sequence of regex-based edits perform inter-word changes that depend on  surrounding contexts and allomorphic flags introduced in the morphological analyzer. 
For example, the edits will transform the sequence 
\<ٱلْحُبُّ> m\%\<هُم>
\textit{hum\%m \"Aal.Hub{\SHADDA}u} `they are love' to 
\<هُمُ الْحُبُّ>
\textit{humu Al.Hub{\SHADDA}u}.
For more details, see Appendix~\ref{appendix:contextual-edits}.



\begin{table*}[t!]
\resizebox{\textwidth}{!}{
\begin{tabular}{|c||c||c|c|c||c||c|}
\hline
\textbf{Error Type} & \textbf{Ratio} & \textbf{Post-context} & \textbf{Input Word} & \textbf{Pre-context} & \textbf{Gold Ref} & \textbf{Best Model} \\ \hline
\multirow{2}{*}{\begin{tabular}[c]{@{}c@{}}Missing\\ Analysis\end{tabular}} & \multirow{2}{*}{55\%} & \multicolumn{1}{r|}{\<.>} &	\<العَرمَرَمِ> &	\<إِذا غِبتَ أَو صَدرُ الخَميسِ> &	\<الْعَرَمْرَمِ> &	\<الْعرمرم> \\
&  &  & \textit{Al{\AYN}armarami} &  & \textit{Al.{\AYN}aram.rami} & \textit{Al.{\AYN}rmrm} \\ \hline
\multirow{2}{*}{\begin{tabular}[c]{@{}c@{}}Tagging\\ Failure\end{tabular}} & \multirow{2}{*}{22\%} & 
\multicolumn{1}{r|}{\<زينَةُ الحَفلِ الَّذي أَنتَ زَينُهُ> } &
\<وَمَن> &	&	\<وَمَنْ> &	\<وَمَنِّ>\\
&&&	\textit{waman} &	&	\textit{waman.} &	\textit{waman{\SHADDA}i} \\ \hline
\multirow{2}{*}{\begin{tabular}[c]{@{}c@{}}Wrong\\ Input\end{tabular}} & \multirow{2}{*}{11\%} & \multicolumn{1}{r|}{\<في البلقاء فاستقدمهم>}
& \<أهلهُ>
& \multicolumn{1}{l||}{\<فخاف على>}
& \<أَهْلِهِ>
& \<أَهْلَهُ> \\
& 
& 
& \textit{Âhlhu}
& 
& \textit{Âah.lihi}
& \textit{Âah.lahu} \\ \hline
\multirow{2}{*}{\begin{tabular}[c]{@{}c@{}}Gold Ref\\ Error\end{tabular}} & \multirow{2}{*}{11\%} & \multicolumn{1}{r|}{\<أيضا من عائلة صفا>}
& \<يُطلب>
& \multicolumn{1}{l||}{\<تمرّ دقائق قبل أن>}
& \<يُطْلَبُ>
& \<يُطْلَبَ> \\
&
&
& \textit{yuTlb}
&
& \textit{yuT.labu}
& \textit{yuT.laba} \\ \hline
\multirow{2}{*}{\begin{tabular}[c]{@{}c@{}}Valid\\ Alternative\end{tabular}} & \multirow{2}{*}{1\%} & \multicolumn{1}{r|}{\<إلا القشور وغير ذاك دفين>}
& \<السّقم>
& \multicolumn{1}{l||}{\<تاللّٰه ما عرف الطّبيب من>}
& \<السَّقَمِ>
& \<السُّقْمِ> \\
&
&
& \textit{Als{\SHADDA}qm}
&
& \textit{Als{\SHADDA}aqami}
& \textit{Als{\SHADDA}uq.mi} \\ \hline
\end{tabular}
}
\caption{Analysis of different error types in order of severity.
Alongside each error type, we include a typical example from the Dev set. The \textit{missing analysis} error example is also classifiable as a \textit{wrong input} due to the extra Fatha on its fifth letter; however, we only consider it here as a missing analysis error since the system could still recover from this wrong input error if an analysis was present. }
\label{tab:error_analysis_examples}
\end{table*}

\section{Evaluation}
\label{eval}
\subsection{Experimental Setup}
We evaluate our enhanced model in various settings to assess the impact of different extensions.  
Regarding the \textbf{ranking algorithm}, we compare the CAMeL Tools baseline (\textbf{CT}) with our extended re-ranking approach (\textbf{CT++}).
Additionally, we conduct an \textbf{Oracle} evaluation to determine the upper limits of both versions by selecting the analysis closest to the gold diacritization for each word.
For \textbf{context modeling}, we examine three scenarios: no modeling (\textbf{None}), modeling as standalone utterances (\textbf{Solo}), and full context (\textbf{Full}).
Regarding the \textbf{morphological analyzer database}, we compare the \textbf{Current} CAMeL Tools database with our extended version (\textbf{Extended}).
We assess performance across the Dev and Test sets of {\MWM}'s
six genres, without using any {\MWM} data for training.
Additionally, we evaluate the full system on \textbf{{\wikinewsmax}} without partial diacritization to demonstrate the added value of our extensions.
Our metric is strict word diacritization matching accuracy.



\subsection{Results and Discussion}
\paragraph{{\MWM} Results} Table \ref{tab:dev_results} shows a result breakdown on {\MWM} dataset.
The results are consistent across genres, and both Dev and Test. Under  All Dev, the morphological database modifications improved accuracy by 21.1\% over the baseline model, and the contextual post-editing improved accuracy further by 10.2\%.
With all modifications applied, we achieve a 39.7\% improvement in total, and we achieve close to 95.1\% accuracy relative to Oracle.

%
%
%



We analyzed all the errors in the Dev set (n=354). As shown in Table \ref{tab:error_analysis_examples}, 55\% are due to missing analyses in the morphological analyzers, and 22\% are due to failures in morphological tagging.  The rest of the errors are either wrong {\widi} input (11\%) or  gold reference error (11\%). There were very  few cases of multiple different valid diacritizations (1\%). These patterns were comparable across genres.

\paragraph{{\wikinewsmax} Results}
Our modified system increases the strict matching accuracy of word diacritization from dediacritized input on original \textbf{{\wikinews}} \cite{Darwish:2017:arabic-diac} from 47.1\% (CAMeL Tools baseline) to 84.5\% (our best system).
The corresponding results on our newly annotated multi-reference \textbf{{\wikinewsmax}} improve from 47.5\% (CAMeL Tools) to 89.7\% (our best system).
64\% of the increase in the best system is due to improved basic gold reference in {\wikinews}, and the rest is from the additional references.
The improvements over the baseline show the added value of our modified components, although still below reported  state-of-the-art systems
\cite{elmallah-etal-2024-arabic-diacritization}; however, a direct comparison is not possible due to differences in training datasets and target diacritization standards.
The increase in accuracy in the {\wikinewsmax} set suggests more work is needed in the space of proper Arabic diacritization evaluation.

\section{Conclusions and Future Work}
\label{conc}


In this paper, we conducted a detailed analysis of {\widi} patterns across six different genres of Arabic text.
We developed {\MWM}, a new annotated dataset designed for evaluating models to exploit {\widi} for maximal diacritization.
Additionally, we extended the \textbf{{\wikinews}} dataset to \textbf{{\wikinewsmax}}, aligning it more closely with our definition of maximal diacritization.
We also enhanced an open-source toolkit, \textbf{CAMeL Tools}, to effectively utilize partial diacritization, resulting in improved performance.

In the future, we plan to enhance the 
morphological analyzer's coverage, expand our research to 
include less explored genres like Classical Arabic texts, 
and investigate other \textit{wild} text signals such as Hamza usage and dialectal spelling variations. 
All of the modifications and extensions will be integrated in a future version of CAMeL Tools.
We also plan to study the effect of our system on downstream applications such as text-to-speech \cite{halabi2016modern,Abdelali:2022:natiq} and Arabic romanization \cite{eryani-habash-2021-automatic}.

\section*{Limitations}
\begin{itemize}
\item  
We acknowledge that our downstream application evaluation focused solely on automatic diacritization, an important task in Arabic NLP with potential benefits for other applications. 

\item  We acknowledge that the observations are limited by the selection of genres and sample sizes we studied.

\item  We acknowledge that the definition of \textit{maximal diacritization}, as with \textit{full diacritization}, is an open question and there may be different views on what is essential or not that we did not include or included, respectively.  

\end{itemize}

\section*{Ethics Statement}
 \begin{itemize}
  
\item  We do not violate any preconditions on the pre-existing resources we use to our knowledge.
\item  The texts we release are either already out of copyright, creative commons, or allowable within fair use laws (short snippets). 
\item All new manual annotations were done by the paper authors and were compensated fairly. 
 \item  Like all NLP models, there is a chance that mistakes could be made by the system leading to unintended consequences. However, in this particular setup and problem, we are not aware of any sources of systematic bias or harm.
\end{itemize}

\section*{Acknowledgements}
We thank the anonymous reviewers for their insightful comments, and Kareem Darwish and Hamdy Mubarak for helpful conversations.

\bibliography{anthology-trim,camel-bib-v3,custom}

\newpage 
\appendix

\section{Experimental Details}
\label{appendix:exp_details}
\paragraph{Computational Resources}
All the 
models were run on a MacBook Pro with a 2.3 GHz Quad-Core Intel Core i7 Processor and 32 GB RAM.
All the experiments were done within $\sim$11 CPU hours.



\section{Well-formedness Check}
\label{appendix:wellfornmedness}

\paragraph{Scope} The well-formedness checking rules presented below 
focus on the validity of use of sequences of diacritics and specific letters.
They do not guarantee that a word is correct linguistically, only that it has plausible maximal diacritization, e.g.,
%
the non-word \<كُكَّاكاً>~\textit{kuk\underline{{\SHADDA}a}Ak\underline{{\FATHATAN}A}} would pass but the  real word \<كُتَّاباً>~\textit{kut\underline{a{\SHADDA}}Ab\underline{A{\FATHATAN}}} would fail. 

\hide{ 

\paragraph{Diacritic Clusters}\\

A valid word-final diacritic cluster is:\\
\begin{centering}
\shaddah? ((\fatha | \fathatan | \kasra | \kasratan | \damma | \dammatan | \skun) | \fatha\alif)
\end{centering}

\footnote{No Alif with (\<ء> ', \<أ> > and \<ة> p)}, \<ٮٌ> $\circ$N and \<ٮٍ> $\circ$K)

A valid word-medial diacritic cluster is:\\
\begin{centering}
\shaddah? ((\fatha | \kasra | \damma) | \fatha\alif) | \skun
\end{centering}
  
A null diacritic can follow weak letters representing elongated vowels:
\<ا>\fatha~\textit{aA},  \<و>\damma~\textit{uw}, and \<ي>\kasra~\textit{iy}.

}

\paragraph{Word Diacritization Patterns}
A word is formed of an optional starting pattern, required recurring middle letter pattern and an optional ending pattern.

    \textbf{Starting patterns} are ordered combinations of:
    \begin{enumerate}
        \setlength{\itemsep}{-1pt}
        \setlength{\topsep}{1em}
        \setlength{\partopsep}{0pt}
        \setlength{\parsep}{0pt}
        \setlength{\parskip}{0pt} 
        
        \item Conjunction: \<وَ>~\textit{wa} or \<فَ>~\textit{fa}
        \item Preposition: \<بِ>~\textit{bi}, \<كَ>~\textit{ka} or \<لِ>~\textit{li}
        \item Definite article: \<اَلْ>~\textit{Aal.} with Moon letters, or
        \shaddah$\circ$\<اَل>
        ~\textit{Aal$\circ${\SHADDA}} with Sun letters ($\circ$)
        \item Alif Wasla (\<ا> A) followed by a contextually dependent short vowel (see \textbf{Context} below)
    \end{enumerate}
    
    The \textbf{recurring middle pattern}, repeated one or more times, is either of the following:
    \begin{enumerate}
        \setlength{\itemsep}{0pt}
        \setlength{\topsep}{1em}
        \setlength{\partopsep}{0pt}
        \setlength{\parsep}{0pt}
        \setlength{\parskip}{0pt}
        \item Any letter (except \<ا> \textit{A}) followed by \\
(\shaddah? ((\fatha | \kasra | \damma) | \fatha\alif) | \skun)~~~\textit{{\SHADDA}?((a|i|u)|a{\'a})|\textbackslash\,.) }
        
        \item A long vowel: 
        \<ا>\fatha~\textit{aA},  \<و>\damma~\textit{uw}, or \<ي>\kasra~\textit{iy}.
    \end{enumerate}
    
    \textbf{Ending pattern} is one of the following:
    \begin{enumerate}
        \setlength{\itemsep}{0pt}
        \setlength{\topsep}{1em}
        \setlength{\partopsep}{0pt}
        \setlength{\parsep}{0pt}
        \setlength{\parskip}{0pt}
        \item A valid Tanwiyn cluster:  any letter (except~\<ا>~\textit{A}) followed by
  \shaddah?(\fathatan | \kasratan | \dammatan)~~~\textit{{\SHADDA}?({\FATHATAN}|{\KASRATAN}|{\DAMMATAN})}. A bare Alif follows \fathatan~\textit{\FATHATAN} for all letters except the following final sequences: \<أ>~\textit{{\AHAMZAUP}}, \<اء>~\textit{A'}, \<ة> {\TAMARBUTA}.

        \item Waw-of-Plurality: \<وا>\damma~\textit{uwA} or \<وْا>\fatha ~\textit{w.A}
    \end{enumerate}

\paragraph{Double Consonants}  Two consecutive letters with a Sukun each or a letter with Sukun followed by a letter with a Shadda are not allowed, unless at the end of the context.

 \paragraph{Exceptions} There is a small number of allowable exceptions, such as the name `{Amr}' ending with silent \<و>~\textit{w}: \<عَمْرَو> \textit{{\AYN}am.raw} and \<و>\<عَمْرٍ> \textit{{\AYN}am.r{\KASRATAN}w}.
 

\paragraph{Context} The rules apply per word, but have access to surrounding context words to validate contextual diacritics. Punctuation marks and sentence boundaries constitute context delimiters. 
We assume that any unhamzated Alif \<ا> \textit{A} at the beginning of an input word is an Alif Wasla.  There are two context diacritization well-formedness  rules:
(i) All words must end with a vowel representation (short or long) if followed by a word starting with an Alif Wasla;  and
(ii) Context-initial Alif Wasla, e.g., after punctuation or at beginning of sentence, must be followed by a short vowel diacritic. 


\section{Ranking and Re-Ranking}
\label{appendix:ranker}

\paragraph{Ranking in Analysis and Disambiguation}
The ranking process consists of sorting the analyses using a ranking tuple for each analysis as a sorting key.
For an analysis $a$, the ranking tuple $\boldsymbol{r_a}$ is defined as $\boldsymbol{r_a} = (M_a, P_a, L_a, W_a)$ used to sort the analyses where $M_a$ is the number of morphological features matching the tagger's prediction, $P_a$ is the joint part-of-speech and lemma log probability of the analysis, $L_a$ is the lemma log probability of the analysis and $W_a$ is diacritized orthography of the analysis.
While sorting analyses, their respective ranking tuples are compared item-wise only comparing subsequent items when the previous items are equal.
$W_a$ is used to sort analyses lexicographically when all the prior items are equal to ensure a stable ranking.

\paragraph{Re-ranking Modifications}

We extend this tuple with four new features.
We first compute the Levenshtein distance between the original input word and $W_a$, both normalized to a UNF (see Footnote \ref{footnote-unf}) with consistent Tanwiyn Alif order.
We count the number of insert operations ($I_a$), substitution operations ($S_a$) and deletion operations ($D_a$).
We then create a new ranking tuple $\boldsymbol{r_a^\prime} = (S_a + D_a, M_a, S_a, D_a, P_a, L_a, I_a, W_a)$ which we use to re-rank the analyses.
This new ranking tuple has the following properties:
\begin{itemize}
    \setlength{\itemsep}{0pt}
        \setlength{\topsep}{1em}
        \setlength{\partopsep}{0pt}
        \setlength{\parsep}{0pt}
        \setlength{\parskip}{0pt} 
\item When an input word contains any diacritic, the analyses that change these diacritics the least are ranked higher.
\item When an input word has no diacritics, $\boldsymbol{r_a}$ and $\boldsymbol{r_a^\prime}$ produce very similar rankings.
\item When an input word is fully diacritized, any exact matching analysis will have the top rank.
\item Deletions are preferred over substitutions since some diacritics can be removed while substitutions change the word meaning. 
\end{itemize}


\onecolumn

\section{Comparison of Different Diacritization Schemes}
\label{appendix:fulldiacs}

\begin{table*}[h!]
    \centering
    \includegraphics[width=\textwidth]{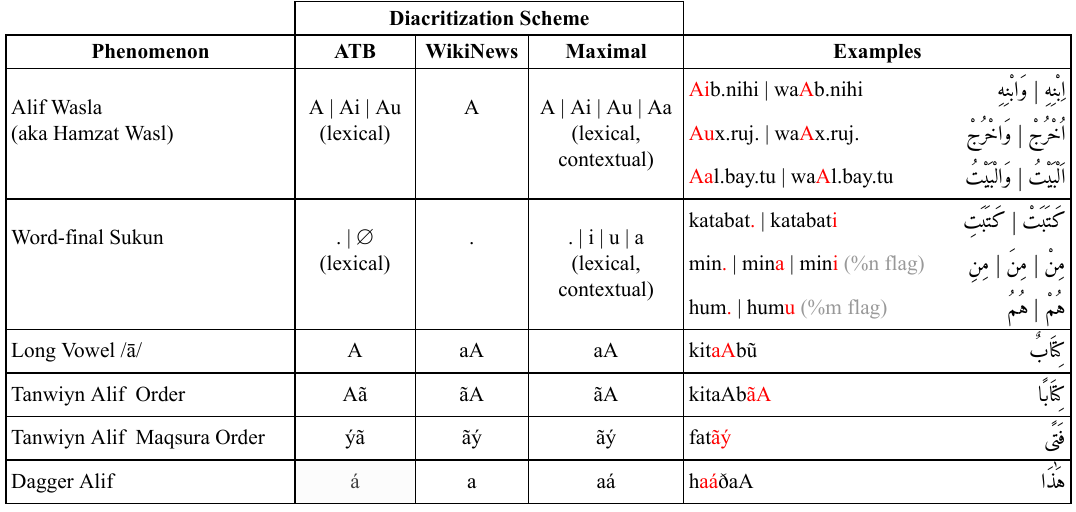}    
    \caption{A comparison of different full diacritization schemes used in the literature.}
    \label{fig:compare-diac-full-types}
\end{table*}

\section{Morphological Analyzer Modifications}
\label{appendix:morph-edits}


\begin{table}[h!]
    \centering
    \includegraphics[width=1.0\textwidth]{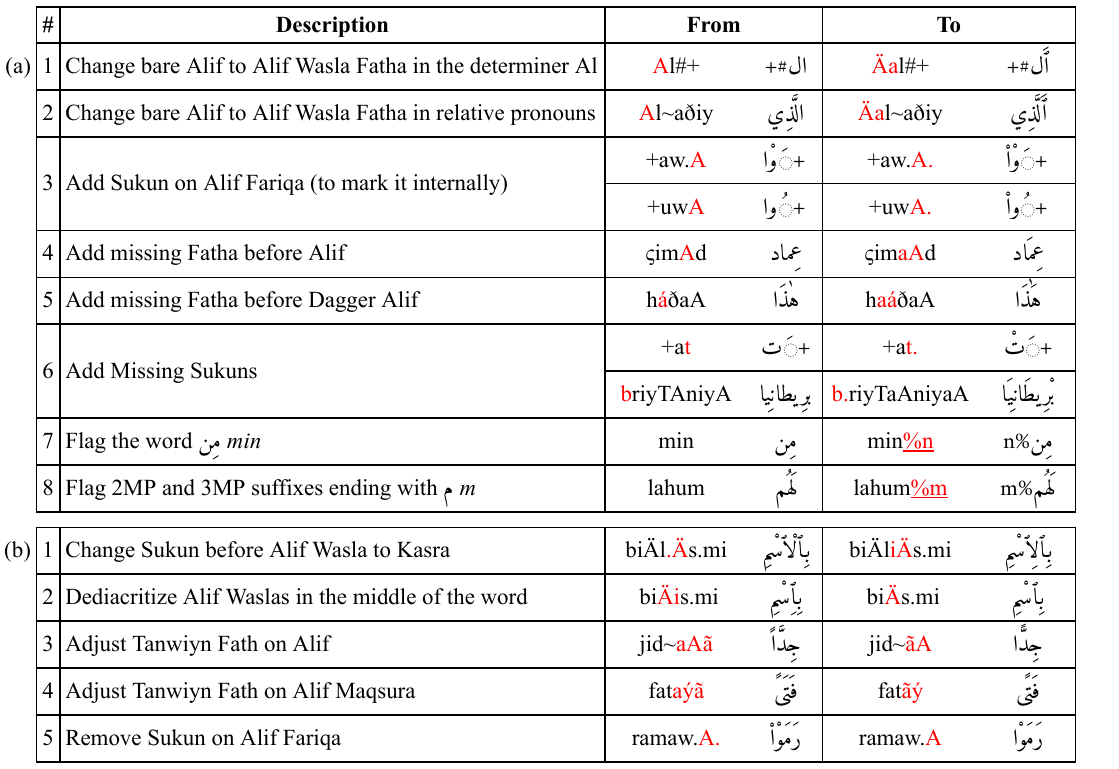}    
    \caption{The most important morphological analyzer modifications with examples: (a) lists database modifications on various prefixes, stems and suffixes; and  (b) lists online edits to analyses after prefix-stem-suffix concatenation.}
    \label{fig:new-analyzer-examples}
\end{table}

\newpage
\section{Contextual Edits}
\label{appendix:contextual-edits}


\begin{table}[h!]
    \centering
    \includegraphics[width=\textwidth]{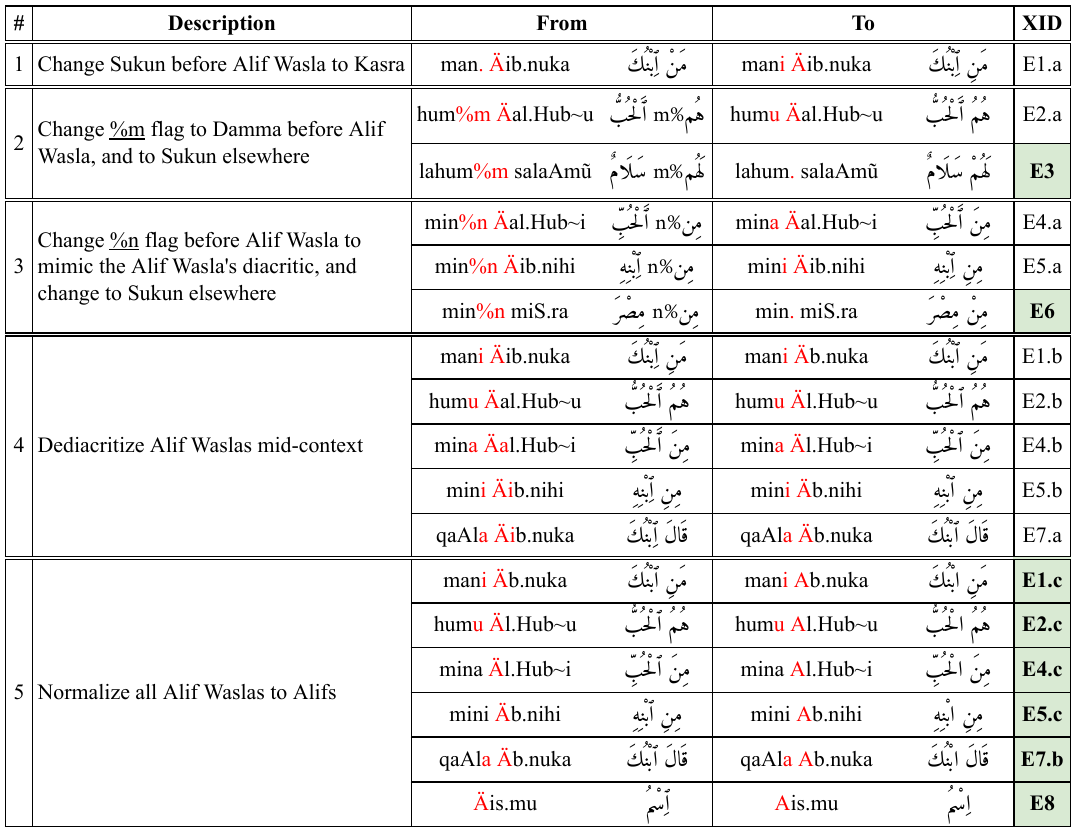}    
    \caption{Our system's contextual edits with examples. XID serves as the running example identifier, demonstrating the changes as rules are successively applied to the example. The green bolded cells mark the final forms.}
    \label{fig:contextual-post-edits-examples}
\end{table}


\hide{
* TODO: inter-annotater agreement

* Question: Why did we choose to do only aaw, abr, hyt, qds

* Question: Do we present statistics for each gigaword subsection separately? Also only for aaw and hyt or for all?

* Question: Inter-annotator agreement?

* Question: Where are we going to talk about unicode normalisation?
}
\hide{
hat are the patterns of Arabic diacritics in the wild?

(0) [out-of-context] distribution of singular diacritics; diacritics by themselves; 

(1) [character-level-context] diacritic clusters within word (valid invalid discussion); 
    (a) orthographic validity within word context of diacritic clusters

(2) [word-level-context] orthographic diacritic completeness; full diacritization test: every letter should have a valid diacritic, if required, should pass the validity of well-formedness;  * maybe leave this out, needs more thinking; word level context
        -> we created the data set

(3) [sentence-level-context] correctness in the context of sentence
    of all diacritized words, how many of them are correct and complete?
    of all diacritized words, how many of them are correct but incomplete?
    of all diacritized words, how many of them are incorrect but complete?
    of all diacritized words, how many of them are incorrect and incomplete?

(*) happens all of them; differences across corpora:
    distribution of diacritic characters
    distribution of the type of diacritics (fully/valid/invalid)
        full diacritics are not common, showing examples of exceptions
    for the invalid cases, what's the distance from the idealized cases?

(3) correct/complete
    of all diacritized words, how many of them are correct and complete?
    of all diacritized words, how many of them are correct but incomplete?
    of all diacritized words, how many of them are incorrect but complete?
    of all diacritized words, how many of them are incorrect and incomplete?
}
\hide{


}

\hide{
\fatha
\fathatan
\kasra
\kasratan
\damma
\dammatan
\skun
\shaddah
}
\hide{
In the wild, there is no mechanism to enforce this ordering constraint, and in fact, we find many nonsensical combinations do occur (Section~\ref{data}).

We define valid full or partial diacritization as
\begin{enumerate}
\setlength{\itemsep}{0pt}%
\setlength{\topsep}{0pt} 
\setlength{\partopsep}{0pt}
\setlength{\parsep}{0pt}
\setlength{\parskip}{0pt}
    \item Shadda
    \item Short vowel or Sukun
    \item Dagger Alif
\end{enumerate}

based on whether phonologically the letter is geminated, followed by a short vowel or silenced, or followed by a long a vowel, respectively.

A \textbf{diac cluster} is a group of diacritics on a single letter, and a diac cluster is said to be \textbf{canonical} if the order of the diacritics follows the order of the previous numbered list, and a maximum of one diacritic per numbered point is present.




Another design decision we took was enforcing Tanwiyn Fath to come before Alif and Alif Maqsura (\<ٮًا> $\circ$FA and \<ٮًى> $\circ$FY), instead of the equally abundant alternatives ($\circ$AF and $\circ$YF) (Section \ref{data}). This simplifies our full diacritization rules due to the reduction in the number of different letters that can appear bare.

A given lexeme can have multiple possible phonologies, each corresponding to one unique full diacritization. We define a full diacritization being \textbf{correct} when it is consistent with the intended phonology. We would later like to automate the check for valid full diacritization and evaluate different datasets (Section~\ref{data}), which we lack the phonological data for, and thus, we define the property of full diacritization being \textbf{well-formed} to be one that is independent of the specific phonology of a word in a certain context, in other words, it is the state of a full diacritization being \textbf{correct} with respect to some phonological pronunciation out there, not necessarily the one intended in the given text. In other words, a full diacritization is well-formed if it doesn't defy the rules of Arabic phonology.

Some ending and starting diacritics can change based on the surrounding orthographic context. We will name these classes of diacritics \textbf{contextual diacritics}, and a diacritization is \textbf{contextually well-formed} if the contextual diacritics in the word (if any) are consistent with the surrounding context. The two examples of contextual diacritics we are handling in this paper are:

\begin{enumerate}
    \setlength{\itemsep}{0pt}
    \setlength{\topsep}{1em}
    \setlength{\partopsep}{0pt}
    \setlength{\parsep}{0pt}
    \setlength{\parskip}{0pt}
    
    \item Short vowels on word-starting Waslas should be removed when in context
    \item Sukun ending words followed by Alif Wasla should be changed to a short-vowel, Kasra, and Fatha\footnote{When \<مِنْ> mino is followed by the definite aricle \<ال> Al} and Damma\footnote{When \<هُمْ> humo, \<أَنْتُمْ> >anotumo, etc then any Alif Wasla} occasionally
\end{enumerate}

We take contexts as separated by punctuation marks and sentence boundaries. We finally include contextual well-formedness as the last requirement for our definition of full diacritization well-formedness.

We define the rules of Arabic phono-orthography that we enforce for a full diacritization to be considered well-formed (these rules exclusively govern diacritics, in other words, we assume that a morphologically valid lexeme is given to start with):

\begin{enumerate}
\setlength{\itemsep}{0pt}%
\setlength{\topsep}{0pt} 
\setlength{\partopsep}{0pt}
\setlength{\parsep}{0pt}
\setlength{\parskip}{0pt}
    \item A word is formed of an optional starting pattern, required recurring middle letter pattern and an optional ending pattern.

    \vspace{0.5em}
    Starting patterns are ordered combinations of:
    \begin{enumerate}
        \setlength{\itemsep}{0pt}
        \setlength{\topsep}{1em}
        \setlength{\partopsep}{0pt}
        \setlength{\parsep}{0pt}
        \setlength{\parskip}{0pt}
        \item Connective (\<وَ> wa) and (\<فَـ> fa)
        \item Prepositional (\<بِـ> bi), (\<كَـ> ka) and (\<لِـ> li)
        \item Definite article (\<اَلْ> Aalo\footnote{Sukun changes to Kasra when followed by Wasla}) or (\<اَلٮّ> Aal$\circ\hspace{-0.3em}\sim$) if followed by a Moon or Sun letter, respectively
        \item Alif Wasla (\<ا> A), with short vowel\footnote{Dropped in context}
    \end{enumerate}
    
    Recurring middle patterns:
    \begin{enumerate}
        \setlength{\itemsep}{0pt}
        \setlength{\topsep}{1em}
        \setlength{\partopsep}{0pt}
        \setlength{\parsep}{0pt}
        \setlength{\parskip}{0pt}
        \item Any letter (except \<ا> A), with a short-vowel/Sukun including diac cluster
        \item An elongated vowel preceded by its corresponding short vowel
    \end{enumerate}
    
    Ending patterns:
    \begin{enumerate}
        \setlength{\itemsep}{0pt}
        \setlength{\topsep}{1em}
        \setlength{\partopsep}{0pt}
        \setlength{\parsep}{0pt}
        \setlength{\parskip}{0pt}
        \item Tanwiyns (\<ٮًا> $\circ$FA\footnote{No Alif with (\<ء> ', \<أ> > and \<ة> p)}, \<ٮٌ> $\circ$N and \<ٮٍ> $\circ$K)
        \item Waw-of-Plurality (\<وا> wA or \<وْا> woA)
    \end{enumerate}
    \item Two consecutive letters with a Sukun each or a Sukun then a Shadda are not allowed, unless at the end of the context
    \item Exceptions include: (\<عَمْرُو> Eamoruw, \<و>\<عَمْرٍ> EamorKw, \<اللهُ> Allhu, \<للهِ> llhi\footnote{Due to ligatures with diacritics middle letters are empty}, etc)
    \item Contextual well-formedness is satisfied
\end{enumerate}

\section{Ranking and Re-Ranking}
\label{appendix:ranker}

\paragraph{Ranking in Analysis and Disambiguation}
The ranking process consists of sorting the analyses using a ranking tuple for each analysis as a sorting key. For an analysis $a$, the ranking tuple $\boldsymbol{r_a}$ is defined as $\boldsymbol{r_a} = (M_a, P_a, L_a, W_a)$ used to sort the analyses where $M_a$ is the number of morphological features matching the \textit{UBD}'s prediction, $P_a$ is the joint part-of-speech and lemma log probability of the analysis, $L_a$ is the lemma log probability of the analysis and $W_a$ is diacritized orthography of the analysis. While sorting analyses, their respective ranking tuples are compared item-wise only comparing subsequent items when the previous items are equal. $W_a$ is used to sort analyses lexicographically when all the prior items are equal to ensure a stable ranking.

\paragraph{Re-ranking Modifications}

We extend this tuple with four new features. We first compute the Levenshtein distance between the original input word and $W_a$, both normalized to a UNF (see Footnote \ref{footnote-unf}) with consistent Tanwiyn Alif order. We count the number of insert operations ($I_a$), substitution operations ($S_a$) and deletion operations ($D_a$). We then create a new ranking tuple $\boldsymbol{r_a^\prime} = (S_a + D_a, M_a, S_a, D_a, P_a, L_a, I_a, W_a)$ which we use to re-rank the analyses.

This new ranking tuple has the following properties:
\begin{itemize}
\item When an input word contains any diacritic, the analyses that change these diacritics the least are ranked higher.
\item When an input word has no diacritics, $\boldsymbol{r_a}$ and $\boldsymbol{r_a^\prime}$ produce very similar rankings.
\item When an input word is fully diacritized, any exact matching analysis will have the top rank.
\item Deletions are preferred over substitutions since some diacritics can be removed while substitutions will change the meaning of a word.
\end{itemize}
}

\end{document}